\documentclass[twoside,11pt]{article}

\usepackage{jmlr2e}
\usepackage{soul,color}
\usepackage{booktabs}

\ShortHeadings{Domain Adaptation for Acute Respiratory Infection Prediction}{Domain Adaptation for Acute Respiratory Infection Prediction}

\firstpageno{1}

\begin{document}

\title{Domain Adaptation for Infection Prediction from Symptoms Based on Data from Different Study Designs and Contexts}

\author{\name Nabeel Abdur Rehman \email nabeel@nyu.edu \\
       New York University\\
       \AND
       \name Maxwell Matthaios Aliapoulios \email maxwell.aliapoulios@nyu.edu \\
       New York University\\
       \AND
       \name Disha Umarwani \email dhu200@nyu.edu \\
       New York University\\
       \AND
       \name Rumi Chunara \email rumi.chunara@nyu.edu \\
       New York University\\
       } 

\maketitle

\begin{abstract}
Acute respiratory infections have epidemic and pandemic potential and thus are being studied worldwide, albeit in many different contexts and study formats. Predicting infection from symptom data is critical, though using symptom data from varied studies in aggregate is challenging because the data is collected in different ways. Accordingly, different symptom profiles could be more predictive in certain studies, or even symptoms of the same name could have different meanings in different contexts. We assess state-of-the-art transfer learning methods for improving prediction of infection from symptom data in multiple types of health care data ranging from clinical, to home-visit as well as crowdsourced studies. We show interesting characteristics regarding six different study types and their feature domains. Further, we demonstrate that it is possible to use data collected from one study to predict infection in another, at close to or better than using a single dataset for prediction on itself. We also investigate in which conditions specific transfer learning and domain adaptation methods may perform better on symptom data. This work has the potential for broad applicability as we show how it is possible to transfer learning from one public health study design to another, and data collected from one study may be used for prediction of labels for another, even collected through different study designs, populations and contexts.
\end{abstract}

\section{Introduction}
Prediction of infection is an important goal in health. Acute Respiratory Infections (ARIs) are a type of illness with a very large global burden, and the potential to spread rapidly and cause epidemics and pandemics. ARI can be difficult to predict and range widely in their severity; over three decades, the number of influenza deaths alone has ranged from 3,000 to 49,000 people \citep{generous2014global}. Accurate and rapid identification of influenza and other ARIs could: reduce the costs associated with unnecessary investigations, facilitate timelier (and hence more effective) use of antiviral drugs for influenza, help prevent the secondary spread of infection, reduce unnecessary use of antibiotics, reduce economic loss due to productivity declines, and shorten hospital stays. 

There is work approaching this problem of predicting infection from symptom data, however the types of data studies are based on, have limited such efforts. For example, healthcare-based ARI surveillance data, such as from doctors offices or emergency rooms only capture the most severe cases (many cases of ARI are mild, and it could take up to two weeks to obtain laboratory-confirmed diagnostic results, so people don't visit a doctor -- however are still infectious during this period and can transmit infection to healthy individuals, increasing the magnitude of an outbreak). Thus, as these clinical studies involve people who are very sick, and are based on a limited sample of people, predicting infection from a new group of people, or from a different context where the types of symptoms could be different, may be challenging. Community-generated and participatory data are becoming more available given pervasive Internet and digital tools for self-reporting such as web surveys and smartphones, and thus provide coverage of and accessibility to a wide range of symptoms and disease severity in the general population  \citep{ginsberg2009detecting,abdurrehmane1501215}. There are also other surveillance approaches that involve tracking based on home-visits with personnel like healthcare workers. 

However prediction of infection from features like symptoms is not a straight-forward problem due to many reasons. The way symptoms are reported can be subjective. Specifically, when a doctor is present and records a ``fever'', this may be different than how an individual reports they have ``fever'' in a self-report situation. Further, there are typically population differences, seasonal differences and different inclusion criteria between studies (e.g. participants must have at least one, or two symptoms, or symptoms of a certain kind). All of these challenges provide interesting computational questions regarding how we may use information from one study applied to different situations given this consistent task, but differing features and feature spaces.

Today there are many studies and systems that track ARIs in different places around the world via only the collection of syndromic data (e.g. using Google Flu Trends \citep{ginsberg2009detecting}, Flu Near You \citep{chunara2013flu}, etc.). Though these studies recognize the importance of verification of infection, doing so is very costly and laborious. This brings to light the question: can we use data from studies of ARI, though they may be collected in different ways, to improve predictive capability in other studies? Doing so would leverage a lot of disparate work, in different contexts and formats, improving clinical prediction tools and allowing for understanding of how symptoms and infection relate in different populations. As it is difficult for any study to be representative of the entire population, such studies across different data would also allow us to gain understanding of how features may be more or less predictive in the groups they encompass, informing precision medicine and precision public health efforts.


Learning from different types of data is an active area of research in computer science. Specifically, the task of domain adaptation is to develop learning algorithms that can be easily ported from one domain to another (say, from newswire to biomedical documents). This problem is particularly interesting in natural language processing because it often involves the situation of having a large collection of labeled data in one ``source'' domain (say, newswire) but truly desire a model that performs well in a second ``target'' domain. To compensate for the degradation in performance due to domain shift, many domain adaptation algorithms have been developed, most of which assume that some labeled examples in the target domain are provided to learn the proper model adaptation. Some of these domain adaptation methods have shown to work well in specific applications like text classification \citep{daume2009frustratingly}, visual domains \citep{fernando2013unsupervised} and sentiment prediction in text domains \citep{blitzer2007biographies}. However, the type of data being used can impose specifics about what potential domains can look like (e.g. the general feature set sizes, and distance between domain spaces). Therefore in this application paper, we apply and investigate domain adaptation in a setting that accurately captures what happens in public health.

There has been an increasing amount of work in domain adaptation, yet we have not seen any applied to public health data or infection prediction.
While much transfer learning and domain adaptation work is methods based, there is a need to assess how such methods perform in different domains and for different data types. For example, how does domain adaptation perform for predicting vaccine sentiment based on movie sentiment information in text data, or prediction of infection for self-reported data from clinical information (this study)? Thus this application is important to understand how domain adaptation can work in such an area and understand how different the feature spaces can be. Therefore, here we implement and evaluate state-of-the art domain adaptation methods to assess if it is possible to combine data from real epidemiological studies collected in different ways, at different time periods and based on different populations to predict acute respiratory infection from symptoms. We also discuss why certain domain adaptation methods may be more suited to this type of feature space, in contrast to other problems dealing with text or visual data. 

This is an application paper which will also be useful in future endeavors combining data across public health studies from different contexts and different populations. Overall, this paper assesses the potential for state-of-the art machine learning methods to be applied to a new type of feature space and task that is ubiquitous across many areas of healthcare, and can be used to address pragmatic medical decisions as well as address health challenges of global significance.






\section{Related Work}
The  power of symptoms for predicting the presence of ARIs has been explored in various settings. Monto et al. presented one of the early studies in this domain \citep{monto2000clinical}. Using data from eight clinical trials, the researchers found that individuals suffering from influenza are more likely to manifest fever, cough and nasal congestion symptoms, as compared to healthy individuals. Another study used data from outpatient clinics, and found cough and fever to be the only two symptoms significantly associated with a positive influenza test \citep{boivin2000predicting}. One other study analyzed the data from clinical trials on symptomatic predictors in children, and found cough and fever to be good predictors of influenza \citep{ohmit2006symptomatic}. This study concluded that while predictions can be made for the age group 5-12 years, additional research is required for predicting the presence of influenza in children of age group 1-4 years. Data from clinical practices in West-Australia was used \citep{thursky2003working} and it was found that a combination of cough, fever and fatigue to be the most predictive of influenza. Another study found cough and fever to be the most common symptoms in inflected individuals during the 2009 H1N1 outbreak in China \citep{cao2009clinical}. A study based on individuals 65 and older, found the combination of cough and fever to be more likely present than absent in influenza patients, though the presence of the symptom combination was only marginally higher in infected individuals as compared to healthy individuals \citep{walsh2002clinical}. 

A main assumption made in these works is, despite the datasets being generated from different hospitals or trials, the studies have assumed them to originate from the same domain with similar underlying distribution of symptoms. Hence, the data sets were analyzed concurrently instead of separately accounting for any biases in the datasets, though we see that different studies find different features to be most predictive. A review of data on symptoms and influenza test results from 6 studies, published between 1966-2004  found no symptom to be at least twice as likely to be present in sick individuals as compared to healthy individuals across all datasets \citep{call2005does}. Most recently, and related to this work, logistic regression, random forest and SVM were assessed for prediction of infection from symptoms in a crowdsourced study (GoViral). It was found that while symptom profiles similar to those in clinical studies were most predictive, the specificity of symptom profiles in this dataset were much lower \cite{ray2017predicting}. From this summary we can make several conclusions regarding work to-date in acute respiratory infection predictors. First, studies in the epidemiology literature have been primarily concerned with examining which symptoms are the most predictive for infection, opposed to assessing the predictive power of those symptoms, instead of trying to predict infections based on symptom features. Therefore this task has not been explored, likely due to challenges in bringing multiple datasets together. We can also conclude that while symptoms can be a good predictor of presence of respiratory infections, the symptom profiles are not uniform across healthcare datasets, and there exist inherent biases in the underlying population and data collection methods which makes is difficult to use a single symptom profile criteria across all datasets.


Learning from solving one problem and applying it to a different but related problem is a common paradigm in machine learning. Transfer learning has been widely used in many applications such as image recognition \citep{oquab2014learning} and natural language processing \citep{daume2009frustratingly}, and hospital health care datasets  \citep{wiens2014study}. Methods for transductive learning vary based on the underlying marginal probability distribution and domain of the source and target datasets. If the source and target datasets belong to the same domain but have different underlying marginal probability distribution, then methods such as covariate shift \citep{shimodaira2000improving} and sample selection bias \citep{zadrozny2004learning} are used to first adjust the distribution between the datasets. If the source and target datasets belong to different but related domain, then domain adaptation methods are used. There has been a wide body of work in domain adaptation methods, most specifically tailored for the required application. Broadly, methods in domain adaptation create a new mapping of features, belonging to the source and target dataset, in either the feature space or in kernel space. More recently, methods in deep learning such as convolution neural networks have been adapted to perform domain adaptation, but these methods require extensive training data to learn representations \citep{oquab2014learning}. The ``Frustratingly Easy Domain Adaptation'' method, is notable for it's simplicity, and good performance on text data \cite{daume2009frustratingly}. 

While words may be used in different types of corpora in different contexts, those roles are discrete and finite. In contrast, to-date, the possible feature space of symptom reports has not been understood. However, the broad approach of domain differences still applies; our task of predicting influenza, from symptom profiles, using labelled data does fit broadly into the idea of supervised transductive learning. The task for all of our datasets can be considered the same (predicting infection from symptoms). As well, the difference in how symptoms are reported and how different symptom combinations can be predictive of infection, are indicative of different domains \citep{pan2010survey}. 

\section{Methods}

\subsection{Problem Formulation}
Data garnered from studies in varying contexts and formats can suffer from differences. For example, the way in which people report symptoms can differ, based on the audience or context. ``Fever'' may mean something different when reporting to a doctor in the emergency department, compared to what one may report as ``fever'' themselves in an online self-report. In other words, the source domain is not the same as the target domain, and furthermore, while a symptom may have the same name in different datasets, it can mean different things depending on the context. This is similar to the situation in natural language processing, where words can take on different meanings that are more likely in different sources (e.g. a word like “monitor” is more likely to be a verb in the Wall Street Journal and more likely to be a noun in a hardware corpus).

$\mathcal{X}$ is a set of features (composed of symptoms and demographics), $\mathcal{Y}$ is the set of possible labels (in our case, a positive or negative viral confirmation). Then, from a probabilistic view, $P(X,Y)$ denotes the true underlying joint distribution of $X$ and $Y$, which is unknown, and will differ for every dataset. For datasets that we use as the ``target'' the true underlying joint distribution is denoted by $P_t(X,Y)$ (target domain, $\mathcal{D}_t$) and $P_s(X,Y)$ for that in the ``source'' domain, $\mathcal{D}_s$. Each domain, $\mathcal{D}$ thus consists of a feature space, $\mathcal{X}$ and a marginal probability distribution $P(X)$. $P_t(Y)$, $P_s(Y)$, $P_t(X)$ and $P_s(X)$ are the true marginal distributions of $Y$ and $X$ in the target and source domains. Similarly, $P_t(X|Y)$, $P_s(X|Y)$, $P_t(Y|X)$ and $P_s(Y|X)$ are the true conditional distributions in each domain. A specific value of $X$ is denoted by $x_i$, and similarly $y_i$ is used to denote the corresponding class label; a pair $(x_i,y_i)$ is a labelled instance. We use $D_{s} = \{(\{x_{s_{i}}^{(j)}\}^{F_{s}}_{j=1},y_{s})\}_{i=1}^{n_{s}}$ to denote the set of labeled instances in the source domain and $D_{t} = \{(\{x_{t_{i,j}}^{(j)}\}^{F_{t}}_{j=1},y_{t})\}_{i=1}^{n_{t}}$ in the target domain, where $F_s$ and $F_{t}$ are number of features in Source and Target dataset respectively and $n_s$ and $n_{t}$ are number of observations in Source and Target datasets, respectively. Then, $x_{i}^{(j)}$ is the value of $j^{th}$ feature in $i^{th}$ observation; $j$ ranges from 1 to $F_{s}$ and $F_{t}$ for the source and target datasets respectively.

Specific to our problem, of note is that the target space ($y_i$ $\epsilon$ $\mathcal{Y}$) is the same for all of our datasets (virus or no virus). This is because each case the specimens are processed using gold-standard laboratory procedures for detection of respiratory infections (polymerase chain reaction \cite{weinberg2004superiority}) and we assume the testing panels are consistent. 

Real-world studies will have many different properties, such as number of observations and class balance. Thus, we implement some standardization steps, for the purposes of comparison, with the understanding that these attributes of the dataset will manifest in real-world situations and can also be used proactively (e.g. use a dataset with a larger number of observations for better transfer of information).

With symptom and infection reports, as symptoms can be different for different people, it is likely that the labels may also not be standard within a dataset (every instance of the same symptom/demographic combination may not correspond to the same label). Thus in order to create standardized labels within each dataaset, we assign a value of ``1'' (infected) to those rows with symptom profiles that are more predictive of infection than not, within the dataset ($P(Y=1|X=x)>P(Y=0|X=x)$). For features, we consider combinations of up to two demographics (gender, age group) and symptoms. This was chosen to reduce computational complexity, but also based on known heuristics of predictive symptom profiles, as described in the Related Work section. We also included two combinations of three symptoms (cough, fever and sore throat) and (cough, fever and muscle ache) based on the Centers for Disease Control definition of influenza-like-illness\footnote{https://www.cdc.gov/flu/about/disease/complications.htm}.

For all transfer learning methods, we use a logistic regression model. As our primary goal is to examine the possibility of transferring information between data sets, we started with a simple and interpretable model. While many empirical studies don't adjust for the size of the data set, as it is understood that increasing the amount of labels can improve performance, we also control and study the effect of the amount of source data included. Initially for all implemented algorithms, we create a standardized ratio of training and test data (4:1). As well, for transfer learning algorithms, we keep this amount of target data and add to it the same amount of source (randomly sampled); the proportion of source to target in the training data is 1:1. Finally, the amount of the source data was varied to examine the effect on prediction performance.


\subsection{Preprocessing}

In order to ensure that any differences we assess are only due to domain differences and not class imbalance (when $P_t(Y) \neq{} P_s(Y)$) or covariate shifts (when $P_t(X) \neq{} P_s(X)$), each unique feature (e.g. symptom combination) is sampled, weighted by it's likelihood to appear in the target versus source dataset \citep{bickel2009discriminative}. To correct class imbalance, we sample data from the source in the ratio between classes as present in the target \citep{zhu2007active}. For all models, only significant features (identified when training and testing a dataset on itself) are used.


\subsection{Transfer Learning Methods}
The first methods used to transfer learning between the datasets assume that while the marginal distribution of the source and target datasets could differ, they belong to the same domain, e.g. $\mathcal{D}_t$ = $\mathcal{D}_s$ (feature spaces are the same). While this assumption might not hold given the datasets collected in different ways (or from different populations), we also implement three domain adaptation methods which allow for differences in the feature space ($\mathcal{D}_t \neq{} \mathcal{D}_s$).
 
\subsubsection{Source Only}
For this implementation, we first identify features which are common between both the source and the target dataset (assuming ``fever'' in one dataset means the same as ``fever'' in another), and using those, train the model only on features from the source dataset that overlap 
($\mathcal{D}_s$ for which $\mathcal{X}_t \cap{} \mathcal{X}_s$). The prediction is then made on the target dataset. This model assumes that the domains of both source and target are the same and any difference present in the dataset are entirely in the distribution of the datasets.

\subsubsection{Union} 
For this implementation, we train on the \emph{union} of the source dataset $D_s$ and target dataset $D_t$ using the overlapping features ($\mathcal{X}_t \cap{} \mathcal{X}_s$). This method also assumes domains of the source and target are the same.



\subsubsection{Frustratingly Easy Domain Adaptation (FEDA)}
We selected this model as it has been recently notable for its extreme simplicity: it merely changes the features by making domain specific and common copies, then trains a supervised classifier on the new features from both domains. As well, it has been demonstrated in situations where there is some labelled training data. The method has shown to perform very well (especially for tasks using text data), yet is “frustratingly easy” to implement \citep{daume2009frustratingly}. The method has been widely used for domain adaptation problems in natural language processing. The method generates a source specific, a target specific and a general version for each feature allowing the features to be optimized for the domain they belong. With the three versions of features generated, the coefficient of all versions are then optimized concurrently using a supervised learning method. For our application we adapt this method by using individual and combination of symptoms as features, instead of using words as features as shown in natural language processing application. Given the variation in symptoms between datasets, we use two variations of this method. We define $x_{ov}$ as set of common features between datasets $D_{s}$ and $D_{t}$, and $F_{ov}$ as the total number of these common features.\\

$x_{ov}(s)$ represents the values of common features copied from source data and $x_{ov}(t)$ represents the values of common features copied from Target dataset. $O_{s}$ = $\langle0, 0, 0.....0\rangle$ of size $F_{s}$. $O_{t}$ = $\langle0, 0, 0.....0\rangle$ of size $F_{t}$.

The augmented Dataset will be: $D_{a} = \{(\{x_{a_{i}}^{(j)}\}^{F_{a}}_{j=1},y_{a})\}_{i=1}^{n_{a}}$ where, $F_{a}$ = $F_{s} + F_{t} + F_{s \cup t}$
and $n_{a}$ = $n_{s} + n_{t}$. Giving:

\begin{equation}
(x_{a_{i}},y_{a}) = 
\left \{
  \begin{tabular}{c c}
   $\big\langle x_{s}, x_{ov}(s), O_{t} \big \rangle, y_{s} $&   $for\ observations\ originating\ from\ D_{s}$\\
   $\big\langle O_{s}, x_{ov}(t), x_{t} \big \rangle, y_{t}$&  $for\ observations\ originating\ from\  D_{t}$\\
  \end{tabular}
\right \}
\end{equation}

\subsubsection{LinInt}
This domain adaptation method works by linearly interpolating the target (only) and source (only) systems. The interpolation parameter is estimated on held-out (training) target data as in \citep{bisazza2011fill}.

\subsubsection{PRED}
This model, also for domain adaptation, uses the source data to build one classifier and then uses this classifier’s predictions as features, for the target data. Then the new target data features are: $\mathcal{X'}_t$ = \lbrack$p_s$;$\mathcal{X}_t$\rbrack{} where $p_s$ are the features from the source classifier (output weights). LinInt and PRED are used as comparisons (baselines) to FEDA \cite{daume2009frustratingly}, in which it is shown that FEDA outperforms both of them for text data.

\begin{table}[h]
  \caption{Dataset descriptions.}
  \label{tab:data}
  \scriptsize
  \begin{tabular}{p{1cm}p{1.5cm}p{1.5cm}p{5cm}p{3.5cm}}
    \toprule
    \cmidrule{1-2}
    Study     & Location     & Num. Observations (positive) & Symptoms Recorded & Design \\
    \midrule
    NYUMC & New York City & 21907 (583) & cough, diarrhea, fatigue, fever, headache, muscle, nausea, sorethroat, vomit &  Clinical (emergency department visits)  \\
    GoViral & New York City & 520 (297) & body aches, chills, cough, diarrhea, fatigue, fever, leg pain, nausea, runnynose, shortness of breath, sorethroat, vomit &  Citizen science - individuals self report via the Internet \\
    Flu Watch & United Kingdom & 915 (498) & fever, cough, sorethropat, runnynose, blockednose, sneeze, diarrhea, muscle, headache, rash, earache, wheezy, chills, joint aches, loss of appetite, fatigue, vomit, nausea & Health-worker facilitated, but individuals report from home \\
    Hong Kong & Hong Kong   & 4954 (1137) & cough, fever, headache, muscle, phlegm, runnynose, sorethroat &  Secondary infections recorded by a health worker who visits the home\\
    Hutterite1     & Alberta, Canada   & 1281 (616) & blockednose, chills, cough, earache, fatigue, fever, headache, muscle, runnynose, sorethroat & Infectious individuals visited by healthcare workers \\
    Hutterite2     & Alberta, Canada   & 1236 (191) & chills, cough, fever, headache, muscle, runnynose, sinus &  Symptomatic and asymptomatic infections recorded by healthcare workers\\
    \bottomrule
  \end{tabular}
\end{table}

\section{Experiments}
\subsection{Data}
\begin{figure}[htbp]
  \centering 
  \includegraphics[width=0.97\linewidth]{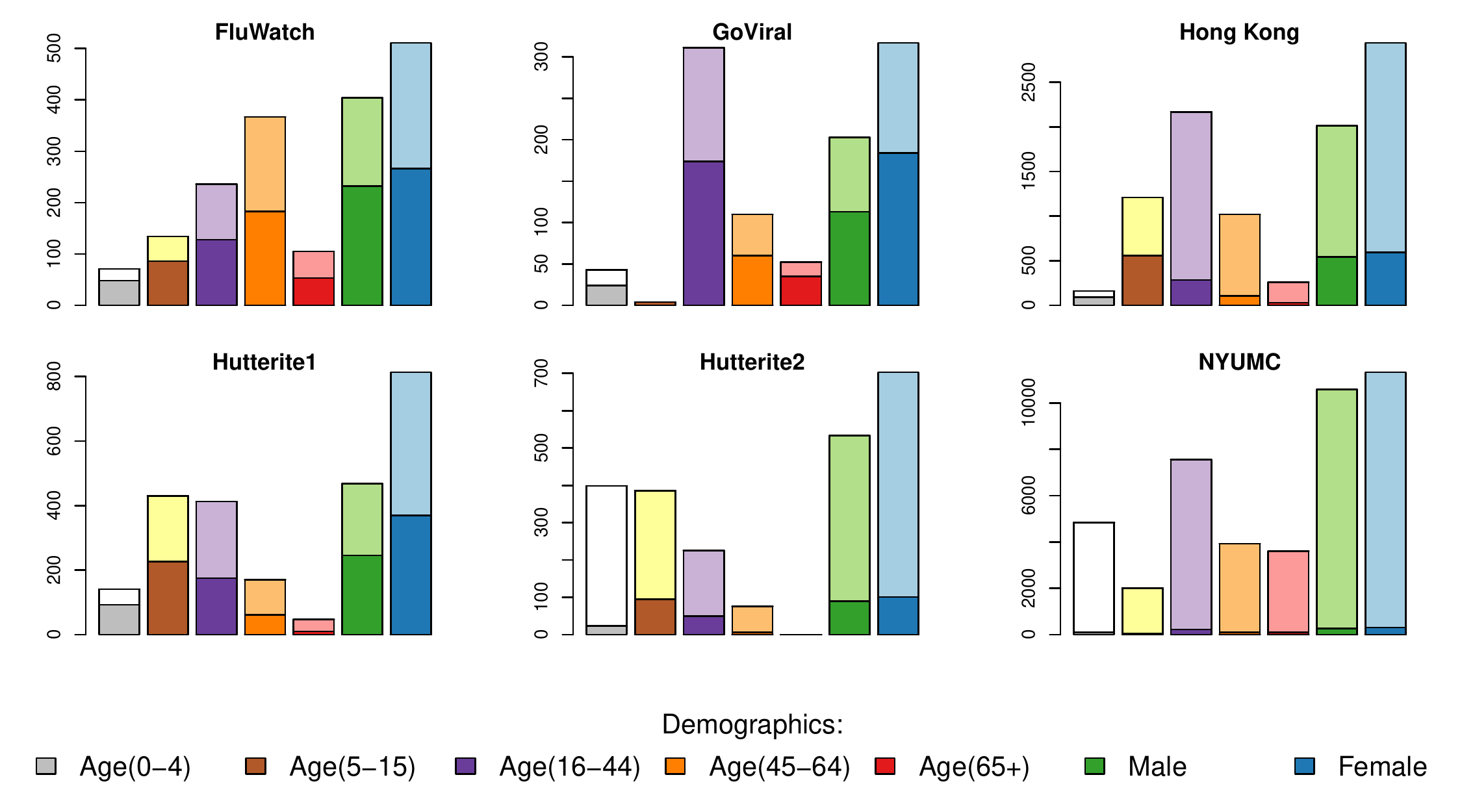} 
  \caption{Breakdown of the positive (darker shade of color) and negative (lighter shade of color) ARI observations across demographics in each dataset.}
  \label{fig:demo} 
\end{figure}

Here we describe the study design and context behind each dataset. Full attributes of the datasets are summarized in Table \ref{tab:data}, while the breakdown of positive and negative observations across demographics is shown in Figure \ref{fig:demo}. Through these differing study designs, one can see how the features may differ based on the stage of illness and included populations (e.g. those who come into the emergency department, versus those are sick but stay at home).

\textbf{NYUMC} data came from outpatients at the NYU Medical Center Emergency Department from Jan 14 2013 - Jan 13 2017. We selected all patients who came in with a chief complaint of an influenza-like-illness using a set of keywords (fever, cough, chills or sorethroat). This is a self-reported complaint that individuals report when they enter the emergency room. Finally, symptoms (from a checklist) are recorded into their record. It should be noted that the chief complaint did not often correlate with what was in the symptom record, so even though we used those keywords to filter the records, this did not decrease representation of other symptoms in the resulting set of records; it was simply used to remove records about unrelated illnesses. The result of an influenza test is also recorded with each record. 

The \textbf{Hong Kong} study was focused on secondary attack rates (defined as the probability that infection occurs among susceptible persons within a reasonable time following known contact with an infectious person). To do so, household contacts of index patients (people who had come to the hospital and were confirmed to be sick) were followed in July and August 2009. These contacts live in close proximity with people who's illness was severe enough to take them to hospital, indicating a high risk for infection. Household members of 99 patients who tested positive for influenza A virus on rapid diagnostic testing were visited at their homes and swabs were collected from the household members by health workers over multiple days and weeks \citep{cowling2010comparative}. 


In the UK, the \textbf{FluWatch} study consisted of households which were recruited from registers of 146 volunteer general practices (GP) across England in seasonal and pandemic influenza over five successive cohorts from 2006–2011 \citep{fragaszy2016cohort}. Individuals participated from their home. In addition to a baseline visit by a nurse, households received participant packs containing paper illness diaries, thermometers and nasal swab kits including instructions on their use and the viral transport medium to be stored in the refrigerator. While participants would generate specimens and illness reports on their own, they were reminded every week via automated phone calls. 

The \textbf{GoViral} platform was developed to evaluate whether a cohort of lay volunteers could, and would find it useful to, contribute self-reported symptoms online and to compare specimen types for self-collected diagnostic information of sufficient quality for respiratory infection surveillance. These participants thus were never visited at home or in person at all. Volunteers were recruited, given a kit (collection materials and customized instructions), instructed to report their symptoms weekly, and when sick with cold or flu-like symptoms, requested to collect specimens (saliva and nasal swab) \citep{goff2015surveillance}. Periodic reminders were sent over email. Data from 2013-2014, 2014-2015, 2015-2016 and 2016-2017 studies is all included. 

Data is also included from two cohort studies that were conducted in \textbf{Hutterite} colonies in Alberta, Canada. The Hutterites are an ethnoreligious group that tend to live together in colonies that are relatively isolated from towns and cities, and therefore are interesting places to examine ARI prevalence given their self-contained nature. The first study (\textbf{Hutterite1}) involved testing for influenza in participants with at least two symptoms. This study extended from December 28, 2008, until June 23, 2009. In the second study (\textbf{Hutterite2}), nasal swabs were collected during 3 influenza seasons (2007–-2008 to 2009–-2010) from both symptomatic and asymptomatic individuals \citep{loeb2012longitudinal}.

\section{Results}
Results for all transfer learning methods, between each data set pair, are illustrated in Figure \ref{fig:results}, and exact results reported in the Appendix. For all  pairs, the domain adaptation methods performed close to the baseline (what we refer to here as the performance of the target dataset on itself), or outperformed the baseline. However, despite outperforming the baseline methods, in most cases of transfer learning (with no domain adaptation), the margin of improvement over the baseline was not large. More specifically, there was varied performance across each of the dataset pairs, which we discuss further here. Amongst the datasets, NYUMC was the most predictive of itself, concluding that the symptom profiles of individuals with similar labels in this dataset was most consistent. Following NYUMC, Hong Kong was the most predictive of itself followed by Hutterite2, Hutterite1, FluWatch and GoViral in that order. Despite having the highest self prediction AUC score (baseline model), NYUMC was the least predictive dataset when using only data from other datasets. Using only NYUMC data to predict other datasets resulted in AUC score of approximately or below 0.5. Thus, though the dataset was self-consistent (symptom profiles that were predictive for those who came into the clinic were similar), this dataset had the most different domain as compared to other datasets used in our study.   

When transferring learning, using a Source only model (which assumes that the domain of both datasets is the same), GoViral and Hong Kong provided the best transfer of learning between each other. The combination of runnynose and cough, and only runnynose in male individuals were both significant and positively correlated with presence of ARI in both datasets (Table \ref{tab:feat} shows the top five significant features from each dataset). This leads us to conclude that the severity and description of these symptoms reported by individuals in both datasets was similar. As well, while demographics in each of the datasets were quite different, overall demographic distributions are somewhat closer in GoViral and Hong Kong (Figure \ref{fig:demo}), with increased representation in the 16-44 age group and a similar male to female ratio). Hutterite1 was the second most predictive of Hong Kong after GoViral data using the Source only model and slightly outperformed GoViral in predicting Hong Kong using the Union model. Similarly Hong Kong was the most predictive of Hutterite1 data in both source only and union model, which assume datasets are from same domain. Overall this shows that symptom reporting in Hong Kong, Hutterite1 and GoViral are most similar to one another, and hence have the closest domains among the datasets used in the analysis.

\begin{table}[h]
  \caption{Top features (p$<$0.05)}
  \label{tab:results}
  \scriptsize
  \centering
  \begin{tabular}{lll}
    \toprule
    \cmidrule{1-2}
   Hong Kong & Hutterite1 & Hutterite2  \\
    \midrule
      age 65+ \& fever& age 0-4 \& runnynose & muscle ache \& runny nose  \\
    
    age 0-4 \& male & cough \& fever  & chills \& fever   \\

     male \& cough & cough \& fever \& muscle ache & cough \& fever  \\
    
     age 0-4 \& fever & cough \& headache & age 16-44 \& male  \\

     age 5-15 \& fever & blocked nose \& fatigue & cough \& fever \& sore throat \\
     \midrule
GoViral & FluWatch & NYUMC \\
\midrule\\
male \& sore throat & age 5-15 \& runnynose & headache \& muscle \\

cough \& runny nose & age 16-44 \& runnynose & fatigue \& muscle ache \\

male \& runny nose & male \& sorethroat &  age 5-15 \& fever\\

runny nose \& sore throat &  age 65+ \& blocked nose &  diarrhea \& muscle ache\\

runny nose \& vomiting & blocked nose \& muscle ache & headache \& sore throat  \\
        \bottomrule
  \end{tabular}
  \label{tab:feat}
\end{table}

We found consistency amongst the remaining datasets; Hutterite2 and FluWatch were the most predictive of each other. Hutterite2 particularly was able to achieve the baseline AUC score of FluWatch dataset using only source data. Therefore, the domain of these two datasets is closest. Prediction of FluWatch data from NYUMC was only possible using the FEDA method. We examined the pipeline and found that this was because after performing class balancing, covariate shift correction, and selection of significant features, only one overlapping feature of NYUMC was left. Given that this specific feature was not more predictive of a positive or negative result, only in the FEDA method which includes the dataset specific features as well, could the model converge.

In terms of methods, the models which account for variation in domains (PRED, LinInt and FEDA), outperformed methods which assume domains of datasets to be the same. The distinction in the performance of methods in particularly visible when predicting datasets which have a high baseline AUC score (good prediction on themselves). As we implemented covariate shifting and class balancing before every method, variation in performance should only reflect the predictive power of datasets and the underlying dynamics of the methods. 

Amongst the domain adaptation methods, all methods performed nearly well and slightly outperformed one another in certain cases. Moreover, for all dataset pairs, the domain adaptation methods performed close to as good as the baseline, or outperformed the baseline method. Despite outperforming the baseline methods, in most cases of transfer learning, the margin of improvement over baseline method was not large. This can be attributed to the fact that unless a dataset is consistent enough to perform well when predicting itself (the same symptom profile always results in a positive or negative virus result), addition of data from other domains will not drastically improve the prediction of the dataset. 

FEDA outperformed other domain adaptation methods across all transfer learning pairs when predicting the Hong Kong data. This can be attributed to the fact that the Hong Kong dataset has close to five thousand observations. Given the fact that the feature space of FEDA is three times the other domain adaptation methods (it includes the source, target and the general version of features), it requires more training examples to learn meaningful coefficients for the features. Thus we find that a relatively larger training sample helps FEDA outperform other domain adaptation methods. This effect is also evident when predicting the GoViral dataset, which has the least number of observations. When predicting GoViral, FEDA underperformed compared to other domain adaptation methods, and in one instance even performed worse than the Source only model. 

The PRED model uses the output of the Source-only model as a feature in training, while LinInt uses the output of the Source only model alongside output from the Target-only model, and then optimizes for prediction of the Target model. 
In some cases when datasets are closer in domain to each other (HK, Hutterite1 and GV, and also FluWatch and Hutterite2), the PRED model performs better, and for those further apart, LinInt performs better, though there are additional metrics such as the number of overlapping features and number of training examples which could impact the performance.


Amongst the methods which assume that the source and target domains are similar, the Union method outperformed the Source only model. This is because of the addition of target data into the model (in the Union model), which allowed the coefficients of the model to be optimized closer to their actual values in the target data. We further examined if changing the balance of source and target data in the Union method had any effect on the performance of the method. We found that increasing the amount of source data while keeping the amount of target data the same during the training consistently decreased performance in all transfer learning pairs. Because the source and target domains are notably different in this application, an increase in the amount of source data forces coefficient of the model to align more with the source domain as compared to the target domain, which decreases predictive capability.

\begin{figure}[htbp]
  \centering 
  \includegraphics[width=0.97\linewidth]{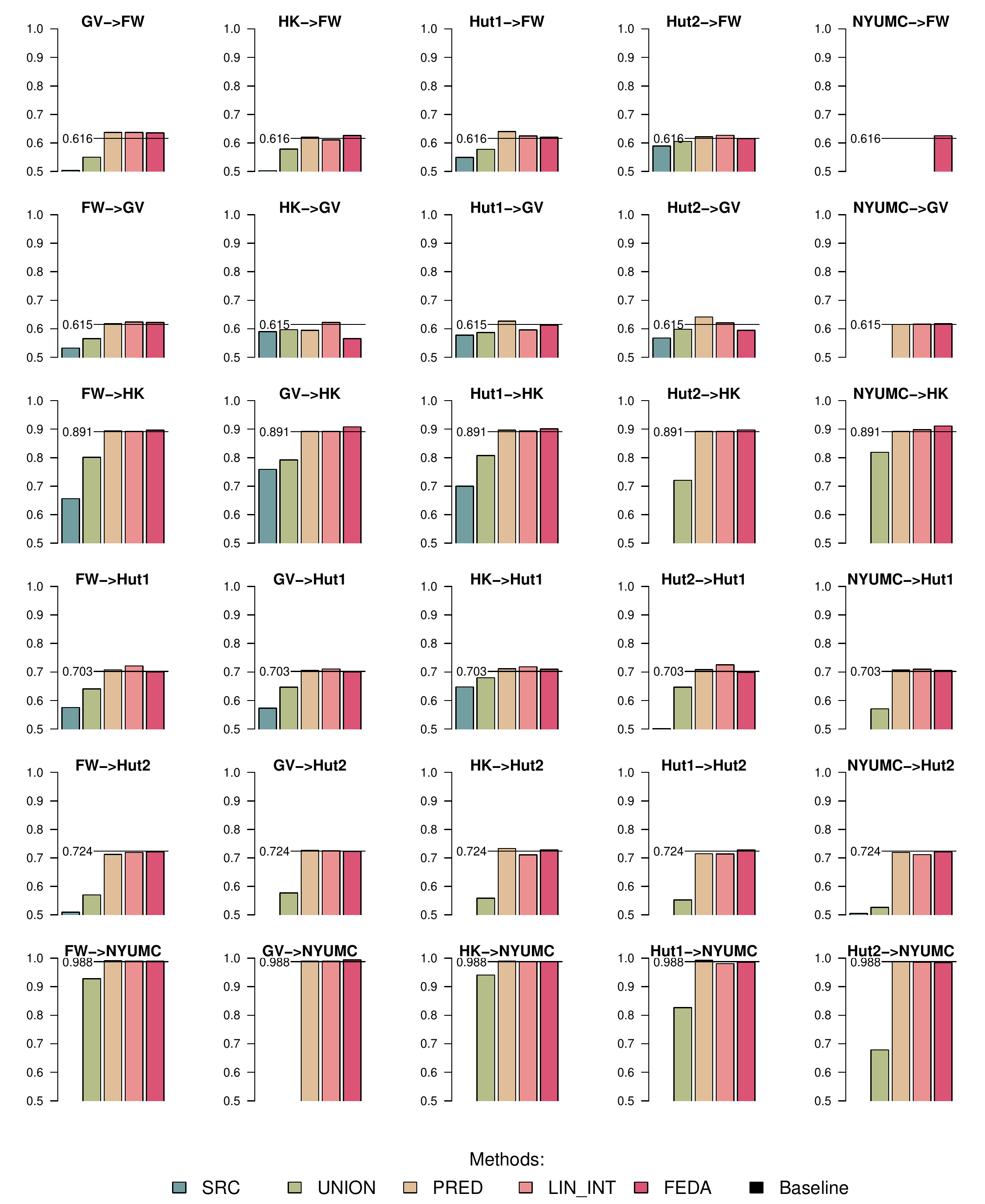} 
  \caption{AUC scores of transfer learning methods across all dataset pairs. Numerical value and stright line represent the AUC score of baseline model. Missing bars represent AUC scores below 0.5.}
  \label{fig:results} 
\end{figure} 

\section{Discussion} 
This study examines the application of machine learning methods to public health efforts of the same task (acute respiratory infection prediction) but from different contexts and study types. We examine the use of transfer learning methods for predicting acute respiratory infection (ARI) from symptoms, understanding that each study will have a different feature space. We considered a wide variety of acute respiratory infection studies (from different contexts, places, and including different populations) all with the same task (acute respiratory infection prediction). This enables us to address even simple questions, such as how different are data sets of symptom/demographic features? And, is it even possible to use data collected from one study to predict data from a different study? 
Our findings reveal interesting characteristics of acute respiratory data collected through different means regarding self-consistency and similarity between data from different studies. We also contrast the performance of several state-of-the-art transfer learning methods, illustrating how these perform in a feature space of ARI symptoms, which to our knowledge has not been examined previously.

This study is important because, although there is a need to perform public health studies in different contexts due to differing reporting practices or disease incidence, we demonstrate how it is possible to use machine learning methods to adjust for domain differences and use these data together. Using domain adaptation allows us to make secondary use of data; optimizing funding and using information garnered for one purpose to gain further understanding. While in public health this is generally not possible (studies may typically only compare data for which the represented groups are exactly the same), this demonstrates a novel and important application of machine learning. Future work should study approaches to include more features in the models (which will increase computational complexity). As well, combining multiple data sets together could be further beneficial, given our positive findings here. Our findings could be integrated into a clinical decision tool to help inform physicians of best use of viral tests in prospective situations where only symptoms are available (as these tests are expensive), or could be used in the strategic design of studies with some labeled data to understand if they may be ``close'' or transferable to other data sets (e.g. have lots of secondary use). As well, we showed that it is possible to transfer information from one type of study to another, which will enable us to predict infection from unlabelled data sets which can be common as viral confirmation is also an expensive procedure and there are many symptom-only data sets that are being collected in public health efforts today. Our findings also identify specific ways transfer learning can be used in this feature space (symptoms for infection prediction). For example, while FEDA has been found to perform well on text data, the feature space sizes here are smaller and thus it doesn't perform as well in general here.

In terms of datasets, this work shows interesting findings regarding the self-consistency and difference between acute respiratory data from different sources. NYUMC was the most self-consistent as it predicted well on itself. Simultaneously, it was the most different from the other datasets (performed poorly when predicting all of the other datasets). On the other hand, GoViral and FluWatch were the least self consistent. These findings may relate to the form of the data collection; NYUMC is the only clinical dataset. It is likely that people who come into the emergency department with ARI all have a similar severity and form of disease. While on the other hand, datasets that were most ``self-collected'' like GoViral and FluWatch (both involved participants self-reporting from home) had the most variance in predictive value of symptom profiles. This aligns with previous work that found symptom profiles are less specific in participatory datasets \citep{ray2017predicting}. These results also reinforce the need for this type of study. While one dataset (e.g. NYUMC) performs well on itself, the data is not very generalizable as it doesn't perform well on other data; instead there is value to assessing predictive performance on other datasets.

Our results also reiterate that although symptoms may mean different things in different situations, the best we can do for prediction requires similarities in those reports (even with domain adaptation, the domains are adapted based on the symptom reports). However domain adaptation methods are still helpful, as they bring entire feature spaces closer. We note this, because amongst the methods which assume that the domains are similar, Union performs better than the Source only model, given that the Union uses part of target data while training the model. This finding, along with the observation that increasing the amount of source data did not improve performance indicates in general to us that there is an appreciable difference in the symptom feature spaces for each of the datasets. As expected, in general when there were common features, prediction worked better. However the use of domain adaptation method resulted in either prediction values close to or better than baseline performance in all transfer learning pairs, indicating that domain adaptation can be used on datasets in this application.

In terms of findings, we also gained an understanding of how different types of domain adaptation methods perform for symptom feature sets. For instance we found that FEDA needs relatively more training data. We also found that LinInt worked best when datasets were ``further'' apart (e.g. had less features in common), which is expected since it interprets between the source and target datasets. We also used domain adaptation methods to adjust for demographic features as well. Datasets which performed better predicting on each other had demographic features that appeared commonly in the most significant features; though a more robust study should examine this how we can use domain adaptation to systematically understand differences in, and transfer information in infection in specific groups (e.g. symptom profiles that are more predictive from certain groups). 



\bibliography{sample}

\appendix
\section*{Appendix}

\renewcommand{\thetable}{A\arabic{table}}
\setcounter{table}{0}

\begin{table}[h]
\centering
\caption{AUC scores of Source only method. A given $[i,j]$ AUC score represents the performance of source data $i$, on target data $j$. }
\label{my-label}
\begin{tabular}{lllllll}
           & FluWatch & GoViral & Hong Kong & Hutterite1 & Hutterite2 & NYUMC \\
FluWatch   & 0.616    & 0.533   & 0.656     & 0.575      & 0.509      & 0.325 \\
GoViral    & 0.503    & 0.615   & 0.759     & 0.573      & 0.444      & 0.429 \\
Hong Kong  & 0.502    & 0.59    & 0.891     & 0.648      & 0.412      & 0.391 \\
Hutterite1 & 0.549    & 0.578   & 0.7       & 0.703      & 0.361      & 0.273 \\
Hutterite2 & 0.589    & 0.568   & 0.432     & 0.501      & 0.724      & 0.462 \\
NYUMC      & NA       & 0.463   & 0.356     & 0.494      & 0.505      & 0.988
\end{tabular}
\end{table}

\begin{table}[h]
\centering
\caption{AUC scores of Union method. A given $[i,j]$ AUC score represents the performance of source data $i$, on target data $j$.}
\label{my-label}
\begin{tabular}{lllllll}
           & FluWatch & GoViral & Hong Kong & Hutterite1 & Hutterite2 & NYUMC \\
FluWatch   & 0.616    & 0.566   & 0.801     & 0.641      & 0.57       & 0.928 \\
GoViral    & 0.55     & 0.615   & 0.792     & 0.647      & 0.577      & 0.48  \\
Hong Kong  & 0.579    & 0.597   & 0.891     & 0.68       & 0.559      & 0.941 \\
Hutterite1 & 0.577    & 0.587   & 0.807     & 0.703      & 0.552      & 0.826 \\
Hutterite2 & 0.605    & 0.598   & 0.721     & 0.647      & 0.724      & 0.678 \\
NYUMC      & NA       & 0.474   & 0.819     & 0.571      & 0.526      & 0.988
\end{tabular}
\end{table}

\begin{table}[]
\centering
\caption{AUC scores of PRED method. A given $[i,j]$ AUC score represents the performance of source data $i$, on target data $j$.}
\label{my-label}
\begin{tabular}{lllllll}
           & FluWatch & GoViral & Hong Kong & Hutterite1 & Hutterite2 & NYUMC \\
FluWatch   & 0.616    & 0.617   & 0.894     & 0.708      & 0.712      & 0.991 \\
GoViral    & 0.636    & 0.615   & 0.892     & 0.706      & 0.727      & 0.99  \\
Hong Kong  & 0.62     & 0.595   & 0.891     & 0.712      & 0.733      & 0.989 \\
Hutterite1 & 0.64     & 0.627   & 0.896     & 0.703      & 0.715      & 0.992 \\
Hutterite2 & 0.621    & 0.642   & 0.893     & 0.708      & 0.724      & 0.987 \\
NYUMC      & NA       & 0.615   & 0.892     & 0.707      & 0.719      & 0.988
\end{tabular}
\end{table}

\begin{table}[]
\centering
\caption{AUC scores of Linear Interpolation method. A given $[i,j]$ AUC score represents the performance of source data $i$, on target data $j$.}
\label{my-label}
\begin{tabular}{lllllll}
           & FluWatch & GoViral & Hong Kong & Hutterite1 & Hutterite2 & NYUMC \\
FluWatch   & 0.616    & 0.623   & 0.892     & 0.721      & 0.719      & 0.99  \\
GoViral    & 0.636    & 0.615   & 0.893     & 0.71       & 0.725      & 0.99  \\
Hong Kong  & 0.61     & 0.622   & 0.891     & 0.718      & 0.71       & 0.988 \\
Hutterite1 & 0.624    & 0.596   & 0.893     & 0.703      & 0.713      & 0.981 \\
Hutterite2 & 0.627    & 0.621   & 0.893     & 0.725      & 0.724      & 0.986 \\
NYUMC      & NA       & 0.616   & 0.898     & 0.71       & 0.712      & 0.988
\end{tabular}
\end{table}

\begin{table}[t]
\centering
\caption{AUC scores of frustratingly easy domain adaptation (FEDA) method. A given $[i,j]$ AUC score represents the performance of source data $i$, on target data $j$.}
\label{my-label}
\begin{tabular}{lllllll}
           & FluWatch & GoViral & Hong Kong & Hutterite1 & Hutterite2 & NYUMC \\
FluWatch   & 0.616    & 0.623   & 0.896     & 0.701      & 0.722      & 0.99  \\
GoViral    & 0.635    & 0.615   & 0.908     & 0.7        & 0.723      & 0.994 \\
Hong Kong  & 0.626    & 0.565   & 0.891     & 0.709      & 0.727      & 0.987 \\
Hutterite1 & 0.62     & 0.613   & 0.901     & 0.703      & 0.728      & 0.985 \\
Hutterite2 & 0.616    & 0.595   & 0.897     & 0.699      & 0.724      & 0.984 \\
NYUMC      & 0.625    & 0.618   & 0.911     & 0.705      & 0.72       & 0.988
\end{tabular}
\end{table}

\end{document}